\documentclass[10pt,twocolumn,letterpaper]{article}

\usepackage{cvpr}              % To produce the CAMERA-READY version
\usepackage{makecell}
\usepackage{multirow}
\usepackage{multicol}

\definecolor{cvprblue}{rgb}{0.21,0.49,0.74}
\usepackage[pagebackref,breaklinks,colorlinks,allcolors=cvprblue]{hyperref}

\usepackage{makecell}
\usepackage{multirow}
\usepackage{multicol}
\usepackage{graphicx}

\usepackage{marvosym}

\def\paperID{*****} % *** Enter the Paper ID here
\def\confYear{2026}

\title{ShotStream: Streaming Multi-Shot Video Generation for Interactive Storytelling}
\vspace{-25pt}
\author{
    Yawen Luo$^{1,\dagger}$\ 
    Xiaoyu Shi$^{2,}$\textsuperscript{\Letter} \ 
    Junhao Zhuang$^1$ \ 
    Yutian Chen$^1$ \ 
    Quande Liu$^2$ \  \\
    Xintao Wang$^2$ \ 
    Pengfei Wan$^2$ \
    Tianfan Xue$^{1,3,}$\textsuperscript{\Letter} \\
    \vspace{2pt}
    $^1$MMLab, CUHK \quad 
    $^2$Kuaishou Technology\quad \\
    $^3$CPII under InnoHK\quad 
    $\textsuperscript{\Letter}$Corresponding author\\
    \textit{\{luoyw0207@gmail.com,\enspace xiaoyushi@link.cuhk.edu.hk, \enspace tfxue@ie.cuhk.edu.hk\}}
    \\
    \vspace{3pt}
    \textcolor{magenta}{\url{https://luo0207.github.io/ShotStream/}}  
    \vspace{2pt}
}

\vspace{10pt}

\begin{document}

\twocolumn[{%
\renewcommand\twocolumn[1][]{#1}%
\maketitle
\vspace{-1.3cm}
\begin{center}
    \centering
    \captionsetup{type=figure}
    \includegraphics[width=\linewidth]{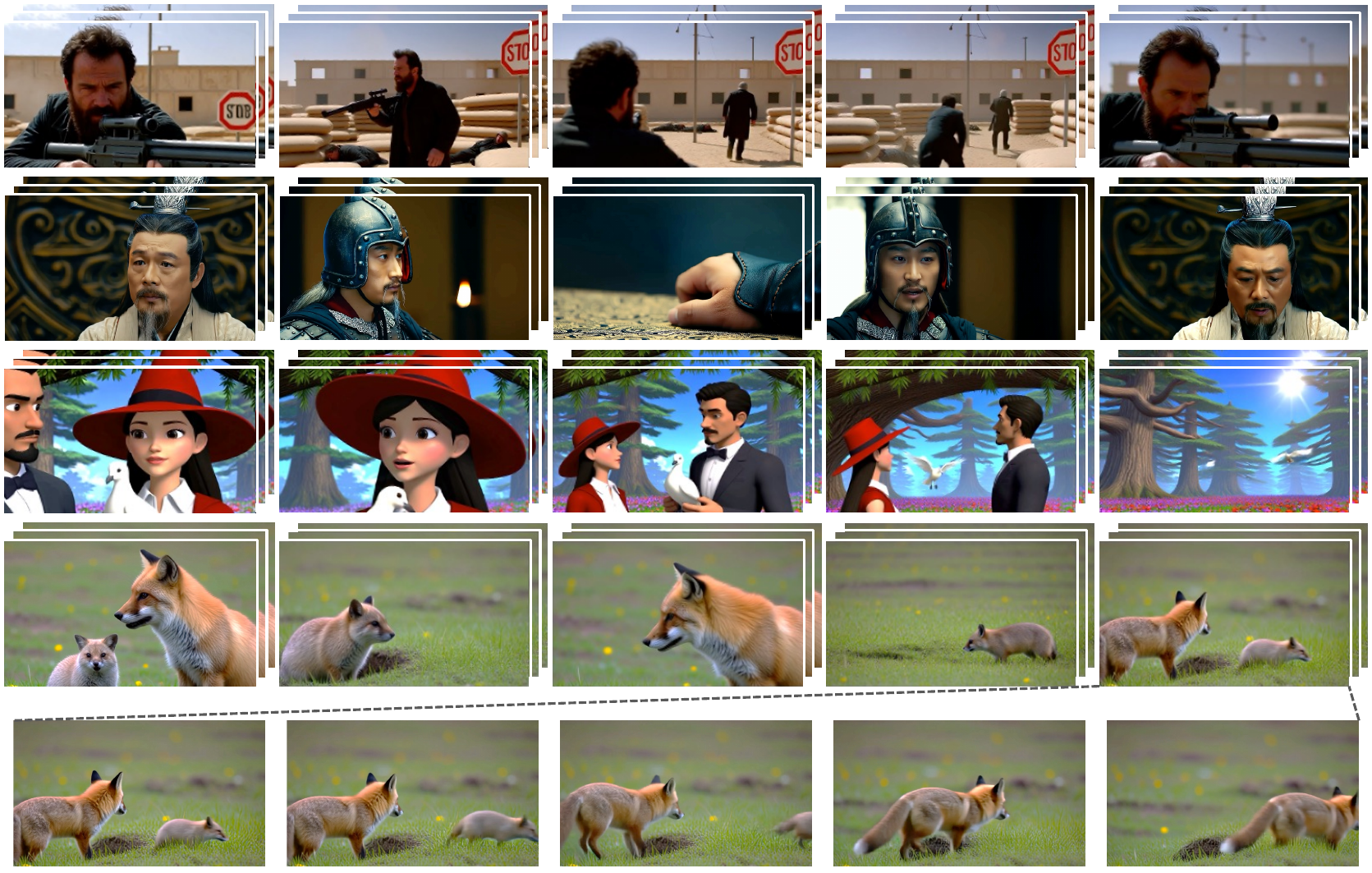}
    \vspace{-15pt}
    \captionof{figure}{Multi-Shot Video results of ShotStream. ShotStream is an autoregressive multi-shot video generation model enabling interactive storytelling and on-the-fly synthesis at 16 FPS on a single GPU. Each case presented here (rows 1–4) illustrates a generated sequence comprising five consecutive shots and 405 total frames, demonstrating the model's capacity to maintain narrative and visual consistency across scene transitions. The expanded bottom row details the high visual quality achieved within a single shot. We highly encourage readers to view our \href{https://luo0207.github.io/ShotStream/}{project page} for the video results.}
    \label{fig:teaser}
\end{center}%
}]

{
    \renewcommand{\thefootnote}{}% 临时清除脚注编号，防止出现 "1 †"
    \footnotetext{$^\dagger$ Work done during internship at  Kling Team, Kuaishou Technology.}
}

\begin{abstract}
Multi-shot video generation is crucial for long narrative storytelling, yet current bidirectional architectures suffer from limited interactivity and high latency. We propose ShotStream, a novel causal multi-shot architecture that enables interactive storytelling and efficient on-the-fly frame generation. By reformulating the task as next-shot generation conditioned on historical context, ShotStream allows users to dynamically instruct ongoing narratives via streaming prompts. We achieve this by first fine-tuning a text-to-video model into a bidirectional next-shot generator, which is then distilled into a causal student via Distribution Matching Distillation. To overcome the challenges of inter-shot consistency and error accumulation inherent in autoregressive generation, we introduce two key innovations. First, a \textit{dual-cache memory mechanism} preserves visual coherence: a \textit{global context cache} retains conditional frames for inter-shot consistency, while a \textit{local context cache} holds generated frames within the current shot for intra-shot consistency. And a RoPE discontinuity indicator is employed to explicitly distinguish the two caches to eliminate ambiguity. Second, to mitigate error accumulation, we propose a two-stage distillation strategy. This begins with intra-shot self-forcing conditioned on ground-truth historical shots and progressively extends to inter-shot self-forcing using self-generated histories, effectively bridging the train-test gap. Extensive experiments demonstrate that ShotStream generates coherent multi-shot videos with sub-second latency, achieving 16 FPS on a single GPU. It matches or exceeds the quality of slower bidirectional models, paving the way for real-time interactive storytelling. Training and inference code, as well as the models, are available on our \href{https://luo0207.github.io/ShotStream/}{project page}.
\end{abstract}    
\section{Introduction}
\label{sec:intro}

While current text-to-video models~\cite{videoworldsimulators2024,wan2025wan,gen3,kling} excel at synthesizing high-fidelity single-shot videos, the field is rapidly advancing toward long-form narrative storytelling~\cite{LCT} akin to traditional film and television. This evolution necessitates multi-shot video generation, which enables the creation of sequential shots that maintain subject and scene consistency while advancing the narrative through varied content. For instance, cinematic techniques like the \textit{shot-reverse shot}~\cite{he2025cut2next} create cohesive interactions by cutting back and forth between characters, effectively guiding the viewer's attention through dynamic perspectives. Driven by the growing demand for such complex cinematic narratives, multi-shot video generation~\cite{an2025onestory,zhang2025storymem,wang2025echoshot,qi2025mask2dit,cai2025mixtureofcontexts,xiao2025captaincinema} has gained increased attention.

Existing multi-shot video generation methods~\cite{LCT,jia2025moga,wang2025echoshot,qi2025mask2dit,cai2025mixtureofcontexts,meng2025holocine,xiao2025captaincinema,wang2025multishotmaster} mainly rely on bidirectional architectures to model intra-shot and inter-shot dependencies, ensuring temporal and narrative consistency. Although effective, these bidirectional architectures suffer from two main limitations: 1) \textit{Lack of interactivity}: Current methods require all prompts upfront to generate the entire multi-shot sequence at once, making it difficult to adjust individual shots without a complete re-generation. A more user-friendly approach would accept streaming prompt inputs at runtime, enabling users to interactively guide the narrative and adapt the current shot based on previously generated content. 2) \textit{High latency}: The computational cost of bidirectional attention grows quadratically with context length, posing a major challenge for long sequences. Even with the integration of sparse attention mechanisms to reduce overhead and accelerate generation, these models still exhibit prohibitive latency. For instance, HoloCine~\cite{meng2025holocine} requires approximately 25 minutes to generate a 240-frame multi-shot video.

\begin{figure*}[!t]
  \centering
   \includegraphics[width=0.9\linewidth]{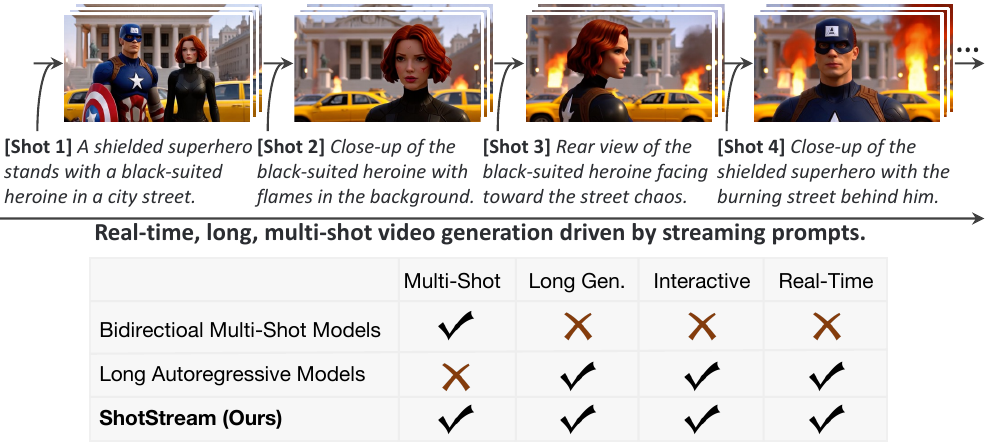}
   \vspace{-6pt}
   \caption{Overview of the ShotStream workflow, which enables real-time, long, multi-shot video generation from streaming prompts.} 
   \vspace{-6pt}
   \label{fig:workflow}
\end{figure*}

To overcome these limitations, we propose ShotStream, a novel causal multi-shot architecture that enables interactive storytelling and efficient on-the-fly frame generation. To achieve interactivity, we reformulate multi-shot synthesis as an autoregressive next-shot generation task, where each subsequent shot is generated by conditioning on previous shots. This reformulation allows ShotStream to accept streaming prompts as inputs and generate videos shot-by-shot, empowering users to dynamically guide the narrative at runtime by adjusting content, altering visual styles, or introducing new characters, as shown in Fig.~\ref{fig:workflow}.
%Furthermore, this next-shot paradigm avoids the heavy training computational costs associated with methods that generate multiple shots simultaneously.

To achieve this efficient causal architecture, we first train a bidirectional teacher model for next-shot prediction, conditioned on historical context. Because past shots comprise hundreds of frames, retaining the entire history introduces severe temporal redundancy and becomes memory-prohibitive. To address this, we condition the model on a sparse subset of historical frames rather than the entire sequence. Specifically, we introduce a dynamic sampling strategy that selects frames based on the number of preceding shots and specific conditional constraints, effectively preserving historical information within a strict frame budget. We then inject these sampled context frames by concatenating their context tokens with noise tokens along the temporal dimension to form a unified input sequence. This concatenation-based injection mechanism is highly parameter-efficient and eliminates the need for additional architectural modules.

Subsequently, we distill this slow, multi-step bidirectional teacher model into an efficient, 4-step causal student model via Distribution Matching Distillation~\cite{yin2024dmd,yin2024dmd2}. However, transitioning to this causal architecture introduces two primary challenges: 1) maintaining consistency across shots, and 2) preventing error accumulation to sustain visual quality during autoregressive generation. To address the first challenge, we introduce a novel \textit{dual-cache memory mechanism}. A \textit{global context cache} stores sparse conditional historical frames to ensure inter-shot consistency, while a \textit{local context cache} retains frames generated within the current shot to preserve intra-shot continuity. Naively combining these caches introduces ambiguity, as the causal model struggles to differentiate between historical and current-shot contexts. To resolve this, we propose a RoPE discontinuity indicator that explicitly distinguishes between the global and local caches. 

The second challenge, error accumulation, stems primarily from the train-test gap~\cite{huang2025self_forcing}. We mitigate this challenge by aligning training with inference through a proposed two-stage progressive distillation strategy. We begin with intra-shot self-forcing conditioned on ground-truth historical shots, where the generator rolls out the current shot causally, chunk-by-chunk, to establish foundational next-shot capabilities. We then progressively transition to inter-shot self-forcing using self-generated histories. In this stage, the model rolls out the multi-shot video shot-by-shot, while generating the internal frames of each individual shot chunk-by-chunk. This strategy bridges the train-test gap and significantly enhances the quality of autoregressive multi-shot generation.

Extensive evaluations demonstrate that ShotStream generates long, narratively coherent multi-shot videos (as shown in Fig.~\ref{fig:teaser}) while achieving an efficient 16 FPS on a single NVIDIA H200 GPU. Quantitatively, our method achieves state-of-the-art performance on the test set regarding visual consistency, prompt adherence, and shot transition control. To complement these metrics with subjective evaluation, we conduct a user study involving 54 participants. Users are asked to compare 24 multi-shot videos generated by our method against those from baselines. The results reveal a decisive user preference for ShotStream in terms of visual consistency, overall visual quality, and prompt adherence.

In summary, our main contributions are as follows:
\begin{itemize}
    \item We present ShotStream, \textit{a novel causal multi-shot long video generation architecture} that enables interactive storytelling and on-the-fly synthesis.
    \item We reformulate multi-shot synthesis as a next-shot generation task to support interactivity, allowing users to dynamically adjust ongoing narratives via streaming prompts.
    \item We design a novel dual-cache memory mechanism for our causal model to ensure both inter-shot and intra-shot consistency, coupled with a RoPE discontinuity indicator to explicitly distinguish between the two caches.
    \item We propose a two-stage distillation strategy that effectively mitigates error accumulation by bridging the gap between training and inference to enable robust, long-horizon multi-shot generation.
\end{itemize}          
\section{Related Work}
\label{sec:related_work}

\begin{figure*}[!t]
  \centering
   \includegraphics[width=1.0\linewidth]{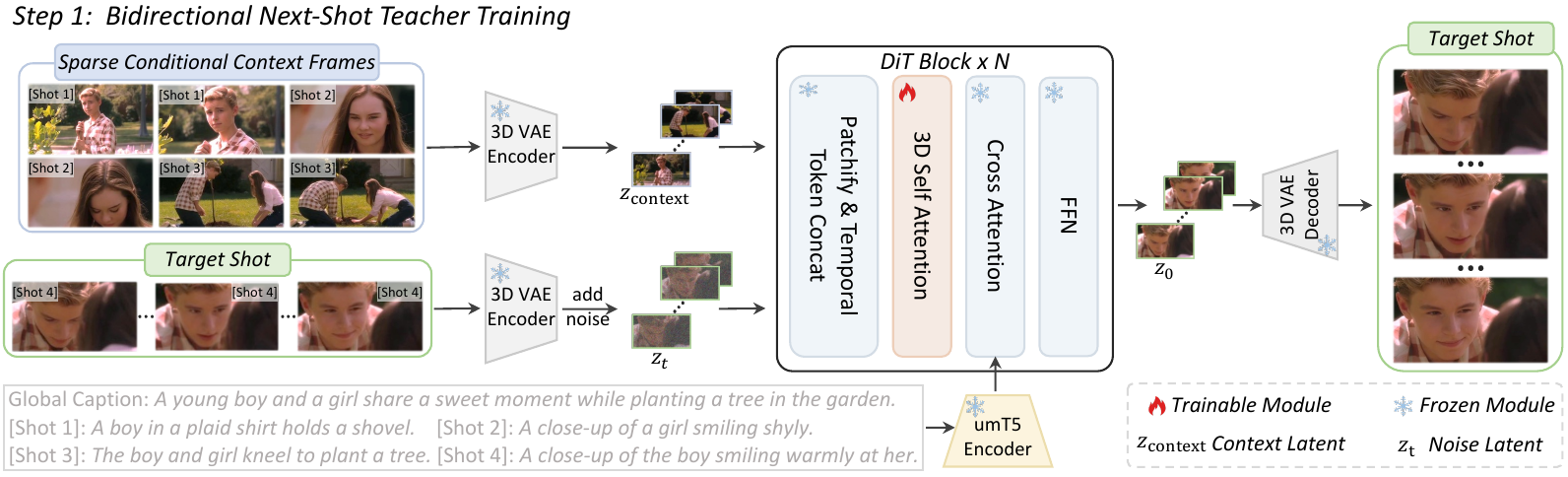}
   \vspace{-15pt}
   \caption{Architecture of the Bidirectional Next-Shot Teacher Model. To realize ShotStream, we first fine-tune a text-to-video model into a bidirectional next-shot model, which generates subsequent shots conditioned on sparse context frames from preceding shots. These conditional context frames are encoded into latents via a 3D VAE and injected by concatenating them with noise latents along the temporal dimension. Notably, only the 3D spatial-temporal attention layers within the DiT Blocks are optimized during fine-tuning. A 4-shot example is shown here for illustration.} 
   \vspace{-12pt}
   \label{fig:method_teacher}
\end{figure*}
\subsection{Multi-Shot Video Generation}
Driven by interest in narrative video generation, multi-shot video synthesis has advanced rapidly~\cite{LCT,wang2025echoshot,meng2025holocine,xiao2025captaincinema,cai2025mixtureofcontexts,jia2025moga,wang2025multishotmaster,qi2025mask2dit,zhou2024storydiffusion}. Current methods generally fall into two categories. Keyframe-based approaches~\cite{zheng2024videogen_of_thought,zhou2024storydiffusion,xiao2025captaincinema} generate the initial frames of each shot and extend them using image-to-video models. However, they often struggle with global coherence, as consistency is enforced only at the keyframe level while intra-shot content remains isolated. The second paradigm, unified sequence modeling~\cite{LCT,wang2025echoshot,qi2025mask2dit,meng2025holocine,jia2025moga,cai2025mixtureofcontexts}, jointly processes all shots within a sequence. For instance, LCT~\cite{LCT} applies full attention across all shots using interleaved 3D position embeddings to distinguish them. While efficiency-focused variants like MoC~\cite{cai2025mixtureofcontexts} and HoloCine~\cite{meng2025holocine} employ dynamic or sparse attention patterns to reduce computational burden, they still suffer from high latency. Furthermore, their bidirectional architectures and unified modeling inherently limit interactivity, complicating the adjustment of specific shots within a generated sequence.

\subsection{Autoregressive Long Video Generation}
Driven by next-token prediction objectives, autoregressive models naturally support the gradual rollout required for long video generation~\cite{bruce2024genie,wang2024loong,weissenborn2019scaling}. Recently, integrating autoregressive modeling with diffusion has emerged as a promising paradigm for causal, high-quality video synthesis~\cite{teng2025magi_1,yang2025longlive,huang2025self_forcing,cui2025self_forcing_++,liu2025rolling_forcing,lu2025reward_forcing,yin2024causalvid,yesiltepe2025infinity,chen2025skyreelsv2,yi2025deep_forcing}. Methods like CausVid~\cite{yin2024causalvid} achieve low-latency streaming by distilling multi-step diffusion into a 4-step causal generator. To mitigate exposure bias from the train-test sequence length discrepancy, Self Forcing~\cite{huang2025self_forcing} and Rolling Forcing~\cite{liu2025rolling_forcing} condition generation on self-generated outputs and progressive noise levels, respectively, to suppress error accumulation. Additionally, LongLive~\cite{yang2025longlive} enables dynamic runtime prompting via a KV-recache mechanism. Despite these advancements, existing techniques are largely confined to single-scene generation and struggle with multi-shot narratives. Our method addresses this gap, extending autoregressive modeling to generate cohesive, multi-shot narrative videos.

\section{Preliminary}
\label{sec:preliminary}

\subsection{Distribution matching distillation}
Distribution Matching Distillation (DMD)~\cite{yin2024dmd,yin2024dmd2} distills slow, multi-step diffusion models into fast, few-step student generators while maintaining high quality. The key objective is to match the student and teacher at the distribution level by minimizing the reverse KL divergence between the smoothed data distribution, $\boldsymbol{p}_{\textrm{data}}$, and the student generator’s output distribution, $\boldsymbol{p}_{\textrm{gen}}$. This optimization is performed across random timesteps $t$, where the gradient is approximated by the difference between two score functions: one trained on the true data distribution and another trained on the student generator's output distribution using a denoising loss. Detailed in the Sec.~\ref{sec:supp_dmd} of the Supplementary Material.

\subsection{Self Forcing}
Error accumulation~\cite{ning2023elucidating,schmidt2019generalization} remains a persistent challenge in autoregressive video generation, caused by the discrepancy between using ground-truth data during training and relying on imperfect predictions during inference. To bridge this train-test gap, self forcing~\cite{huang2025self_forcing} introduces a training paradigm that explicitly unrolls the autoregressive process. By conditioning each frame on previously generated outputs rather than ground-truth frames, the model is compelled to navigate and recover from its own inaccuracies. Consequently, self forcing~\cite{huang2025self_forcing} effectively mitigates exposure bias and stabilizes long generation.
\section{Method}

\begin{figure*}[!t]
  \centering
   \includegraphics[width=0.95\linewidth]{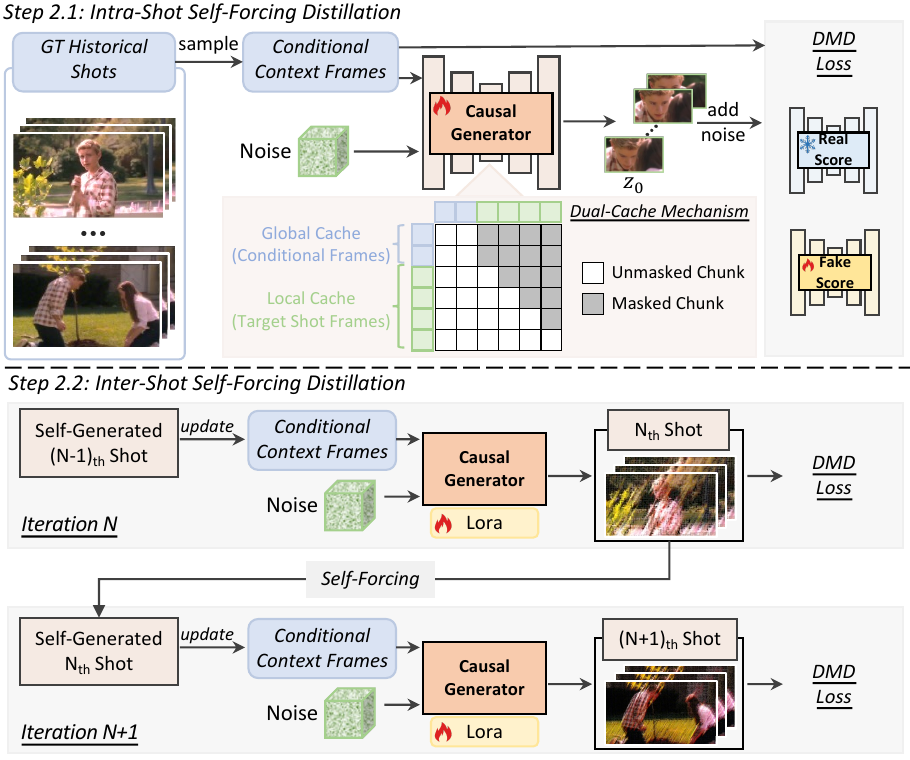}
   \vspace{-5pt}
   \caption{Causal Architecture and Two-Stage Distillation Pipeline. We distill a slow, multi-step bidirectional teacher into an efficient, few-step causal generator. To maintain visual coherence, we propose a novel dual-cache memory mechanism: a global context cache stores conditional frames to ensure inter-shot consistency, while a local context cache retains generated frames within the target shot to guarantee intra-shot consistency. To prevent error accumulation, we employ a progressive two-stage distillation strategy. In the first stage, intra-shot self-forcing distillation (Step 2.1), the model is conditioned on ground-truth historical shots to causally generate the current shot chunk-by-chunk. In the second stage, inter-shot self-forcing distillation (Step 2.2), the model is conditioned on its own previously generated shots, rolling out the video shot-by-shot while iteratively generating the frames of each individual shot chunk-by-chunk.} 
    \vspace{-10pt}
   \label{fig:method_causal}
\end{figure*}

This section details the architecture and training methodology of ShotStream. We first fine-tune a text-to-video model into a bidirectional next-shot model (Sec.~\ref{subsection:next_shot_teacher}). This model is subsequently distilled into an efficient, 4-step causal model via Distribution Matching Distillation. We also propose a novel dual-cache memory mechanism and a two-stage distillation strategy to enable efficient, robust, and long-horizon multi-shot generation (Sec.~\ref{subsection:causal_distill}).

\subsection{Bidirectional Next-Shot Teacher Model}
\label{subsection:next_shot_teacher}
The objective of the next-shot model is to generate a subsequent shot conditioned on historical shots. Since historical shots contain hundreds of frames with high visual redundancy, retaining the entire history is unnecessary and impractical with a limited conditional budget. Therefore, we condition the model on sparse context frames extracted via a dynamic sampling strategy. Specifically, given $S_{\textrm{hist}}$ historical shots and a maximum conditional context budget of $f_\textrm{context}$ frames, we sample $\lfloor f_{\textrm{context}} /S_{\textrm{hist}}\rfloor$ frames from each historical shot, where $\lfloor\cdot\rfloor$ denotes the floor function. Any remaining budget is allocated to the most recent shot to fully utilize the budget, which is set to 6 frames in our experiments.

To condition the model on sampled sparse context frames $V_\textrm{context}$, we employ a temporal token concatenation mechanism, an injection technique proven effective across multi-control generation~\cite{ju2025fulldit}, editing~\cite{ye2025unic}, and camera motion cloning~\cite{luo2025camclonemaster}. Although effective, these methods do not distinguish between the captions of condition frames and target frames; instead, they uniformly apply the target frame's caption to the condition frames. Directly adopting this approach for next-shot generation is problematic, as the captions of previous shots contain crucial information that binds past visual information to textual descriptions. This binding facilitates the extraction of necessary context for generating the subsequent shot. Therefore, we also inject the specific captions corresponding to each conditional context frame into the model, \textit{i.e.}, the frames of each shot attend to both the global caption and the corresponding local shot caption via cross-attention. Specifically, as shown in Fig.~\ref{fig:method_teacher}, our next-shot model reuses the 3D VAE $\varepsilon$ from the base model to transform $V_\textrm{{context}}$ into conditioning latents,

\begin{equation}
    z_\textrm{context} = \varepsilon(V_\textrm{context})\,,
\end{equation}
where $z_\textrm{context} \in \mathbb{R}^{f_\textrm{context}\times c\times h \times w}$ comprises $f_\textrm{context}$ frames, $c$ channels, and a spatial resolution of $h \times w$. Building upon this shared latent space, we first patchify the condition latent $z_{\textrm{context}}$ and the noisy target latent $z_{t}\in \mathbb{R}^{f\times c\times h \times w}$ with $f$ frames into tokens:

\begin{equation}
    x_{j} = \text{Patchify}(z_{j}),z_j\in \{z_{\textrm{context}},z_t\}\,.
\end{equation}
The resulting condition tokens $x_\textrm{{context}}$ and noisy video tokens $x_t$ are then concatenated along the frame dimension to form the input for the DiT blocks:
\begin{equation}
    x_\textrm{{input}} = \textrm{FrameConcat}(x_\textrm{{context}}, x_t)\,.
\end{equation}
The notation $\textrm{FrameConcat}$ denotes that the condition tokens are concatenated with the noise tokens along the frame dimension. Given that the token sequences $x_\textrm{context} \in \mathbb{R}^{b\times f_\textrm{context}\times s \times d}$ and $x_\textrm{t} \in \mathbb{R}^{b\times f\times s \times d}$ share the same batch size $b$, spatial token count $s$ of each frame, and feature dimension $d$, this temporal concatenation yields a combined tensor \(x_\textrm{input}\in\mathbb{R}^{b{\times}(f_\textrm{context}+f){\times}s{\times}d}\). During the training process, noise is added exclusively to the target video tokens, keeping the context tokens clean. This design enables the DiT's native 3D self-attention layers to directly model interactions between the condition and noise tokens, without introducing new layers or parameters to the base model.

\subsection{Causal Architecture and Distillation}
\label{subsection:causal_distill}
The bidirectional next-shot teacher model (detailed in Sec.~\ref{subsection:next_shot_teacher}) requires approximately 50 denoising steps, resulting in high inference latency. To enable low-latency generation, we distill this multi-step teacher into an efficient 4-step causal generator. However, transitioning to this causal architecture introduces two primary challenges: 1) maintaining consistency across shots, and 2) preventing error accumulation to sustain visual quality during autoregressive generation. To address these issues, we propose two key innovations: a dual-cache memory mechanism and a two-stage distillation strategy, respectively.

\noindent \textbf{Dual-Cache Memory Mechanism.} To maintain visual coherence, we introduce a novel dual-cache memory mechanism (Fig.~\ref{fig:method_causal}): a global cache stores sparse conditional frames to preserve inter-shot consistency, while a local cache retains recently generated frames to ensure intra-shot consistency. However, querying both caches simultaneously within our chunk-wise causal architecture introduces temporal ambiguity, as the model struggles to distinguish historical from current-shot contexts. To address this, we propose a discontinuous RoPE strategy that explicitly decouples the global and local contexts by introducing a discrete temporal jump at each shot boundary. Specifically, for the $t$-th latent $z_t$ within the $k$-th shot, its temporal rotation angle is formulated as $\Theta_t = \phi t + k \theta$, where $\phi$ denotes the base temporal frequency and $\theta$ serves as the phase shift representing the shot-boundary discontinuity.

\noindent \textbf{Two-stage Distillation Strategy.} A major challenge in autoregressive multi-shot video generation is error accumulation caused by the training-inference gap~\cite{huang2025self_forcing}. To mitigate this, we propose a two-stage distillation training strategy. In the first stage, \textbf{intra-shot self-forcing} (Fig.~\ref{fig:method_causal}, Step 2.1), the model samples global context frames from ground-truth historical shots while the chunk-wise causal generator produces the target shot via a temporal autoregressive rollout. Specifically, the local cache utilizes previously self-generated chunks from the current target shot rather than ground-truth data. Although this stage establishes foundational next-shot generation capabilities, a training-inference gap remains: during inference, the model must condition on its own potentially imperfect historical shots instead of the ground truth. To bridge this gap, we introduce the second stage: \textbf{inter-shot self-forcing} (Fig.~\ref{fig:method_causal}, Step 2.2). Specifically, the causal model generates the initial shot from scratch and applies DMD. For all subsequent iterations, the generator synthesizes the next shot conditioned entirely on prior self-generated shots. During each iteration, the model continues to employ intra-shot self-forcing to generate each new shot chunk by chunk, applying DMD exclusively to the newly generated shot. This autoregressive unrolling continues until the entire multi-shot video is generated. By closely mirroring the inference-time rollout, this stage aligns training and inference, effectively mitigating error accumulation and enhancing overall visual quality.
 
\noindent\textbf{Inference.}
The inference procedure of ShotStream identically aligns with its training process. ShotStream generates multi-shot videos in a shot-by-shot manner. As each new shot is generated, the global context frames are updated by sampling from previously synthesized historical shots. Within the current shot, video frames are generated sequentially chunk by chunk, leveraging our causal few-step generator and KV caching to ensure computational efficiency.
\section{Experiments}
\label{sec:experiments}

\setlength\tabcolsep{2.5pt} 
\begin{table*}[!t]
\centering
\caption{Quantitative results for multi-shot video generation. The best results are highlighted in \textbf{boldface}, while the second best are \underline{underlined}. Here, Sub., Bg., Cons., and Align. denote Subject, Background, Consistency, and Alignment, respectively.}
\vspace{-6pt}
\label{table:quantitative_results}
\renewcommand{\arraystretch}{1.3}

\resizebox{\linewidth}{!}{
    \begin{tabular}{lcc|cc|ccc|cccc}
        \toprule
        \multirow{2}{*}[-2pt]{Method} & 
        \multirow{2}{*}[-2pt]{Architecture} & 
        \multirow{2}{*}[-2pt]{FPS} & 
        \multicolumn{2}{c|}{Intra-shot Cons.} & 
        \multicolumn{3}{c|}{Inter-shot Cons.} & 
        \multirow{2}{*}[-2pt]{\makecell[c]{Trans. \\ Control }$\uparrow$} & 
        \multirow{2}{*}[-2pt]{\makecell[c]{Text \\ Align. }$\uparrow$} & 
        \multirow{2}{*}[-2pt]{\makecell[c]{Aesthetic \\ Quality }$\uparrow$} & 
        \multirow{2}{*}[-2pt]{\makecell[c]{Dynamic \\ Degrees }$\uparrow$} \\ 
        
        \cmidrule(lr){4-8}
        
        & & & Sub. $\uparrow$ & Bg. $\uparrow$ & Semantic $\uparrow$ & Sub. $\uparrow$ & Bg. $\uparrow$ & & & & \\
        \midrule
        Mask2DiT~\cite{qi2025mask2dit} & Bidirectional & 0.149 & 0.646 & 0.679 & 0.711 & \underline{0.612} & 0.534 & 0.513 & \underline{0.184} & 0.520 &  48.91 \\
        EchoShot~\cite{wang2025echoshot} & Bidirectional & 0.643 & 0.772 & 0.739 & 0.596 & 0.392 & 0.396 & 0.664 & 0.186 & 0.543 & \textbf{65.92} \\
        CineTrans~\cite{wu2025cinetrans} & Bidirectional & 0.413 & \underline{0.776} & \underline{0.797} & 0.459 & 0.412 & 0.459 & 0.572 & 0.170  & 0.513 & 59.47 \\
        \midrule
        Self Forcing~\cite{huang2025self_forcing} & Causal  & 16.36 & 0.737 & 0.707 &  0.738 & 0.542 & 0.445 & 0.633 &0.214 & 0.512& 55.45\\
        LongLive~\cite{yang2025longlive} & Causal  & 16.55 & 0.758 & 0.792 &  0.722 & 0.594 & \underline{0.565} & \underline{0.693} & 0.216 & \underline{0.565} & 58.45 \\
        Rolling Forcing~\cite{liu2025rolling_forcing} & Causal & 15.32 & 0.725 & 0.781 & \underline{0.758} & 0.561 & 0.473 & 0.684 & 0.223 & 0.523 & 62.26 \\
        Infinity-RoPE~\cite{yesiltepe2025infinity} & Causal & 16.37 & 0.752 & 0.738 & 0.622 & 0.453  & 0.407  & 0.715 & 0.209 & 0.513  & 63.40  \\
        \midrule
        ShotStream (Ours) & Causal & 15.95 & \textbf{0.825} & \textbf{0.819} &\textbf{0.762} & \textbf{0.654} & \textbf{0.645} & \textbf{0.978} & \textbf{0.234} & \textbf{0.571} &  \underline{63.56} \\
        \bottomrule
    \end{tabular}
}
% \vspace{-6pt}
\end{table*}

\subsection{Experiment Setup}
\label{sec:experiment_setup}
\noindent\textbf{Implement Details.}
We build ShotStream upon Wan2.1-T2V-1.3B~\cite{wan2025wan} to generate \(832\times 480\) video clips. The bidirectional next-shot teacher is trained on an internal dataset of 320K multi-shot videos. For causal adaptation, the student model is initialized via regression on 5K teacher-sampled ODE solution pairs~\cite{yin2024causalvid}. Distillation proceeds in two stages: intra-shot self-forcing using ground-truth historical shots from the dataset, followed by inter-shot self-forcing using captions from a subset of 5-shot videos. Architecturally, the model operates with a chunk size of 3 latent frames, utilizing a global cache of 2 chunks and a local cache of 7 chunks. We refer readers to the Sec.~\ref{sec:supp_training_details} in the Supplementary Material for further details.

\noindent \textbf{Evaluation Set.} 
To comprehensively evaluate multi-shot video generation capabilities, following previous work~\cite{meng2025holocine,wang2025multishotmaster,wu2025cinetrans,wang2025echoshot}, we leverage Gemini 2.5 Pro~\cite{comanici2025gemini} to generate 100 diverse multi-shot video prompts. To ensure a fair comparison, we tailor these text prompts to match the specific input style of each baseline model. These test prompts cover a wide range of themes, enabling a robust measurement of the models' ability to maintain consistency across different scenes.

\noindent \textbf{Evaluation Metrics.}
Before computing metrics, we use the pretrained TransNet V2~\cite{soucek2024transnet} to detect shot boundaries in each video. We evaluate the model's multi-shot performance across five key dimensions: 1) \textit{Intra-Shot Consistency}: Following HoloCine~\cite{meng2025holocine}, we compute Subject Consistency using DINO~\cite{caron2021dino} cosine similarities and Background Consistency using CLIP~\cite{radford2021clip} similarities across frames. 2) \textit{Inter-Shot Consistency}: Following MultiShotMaster~\cite{wang2025multishotmaster}, we isolate subjects and backgrounds from keyframes using YOLOv11~\cite{khanam2024yolov11} and SAM~\cite{kirillov2023sam}, then assess consistency via DINOv2~\cite{oquab2023dinov2}. Semantic Consistency is measured by computing the cosine similarity of ViCLIP~\cite{wang2023viclip} features extracted from each shot. 3) \textit{Transition Control}: We utilize the Shot Cut Accuracy (SCA) metric~\cite{meng2025holocine} to evaluate the model’s transition control capabilities by measuring the accuracy of cut counts and their temporal precision. 4) \textit{Prompt Following}: We use Text Alignment~\cite{wang2025multishotmaster,wang2025cinemaster} to measure video-text consistency. 5) \textit{Overall Quality}: We report Aesthetic Quality and Dynamic Degrees from VBench~\cite{huang2024vbench} to assess visual quality.

\noindent \textbf{Baselines.}
We compare our model with relevant open-source video generation models of similar scale, including two categories: 1) \textit{Bidirectional Multi-Shot Video Generation Model}: Mask2DiT~\cite{qi2025mask2dit} employs symmetric binary masks within the MMDiT architecture to isolate text annotations per segment while maintaining temporal coherence for multi-scene generation. EchoShot~\cite{wang2025echoshot} targets portrait customization by using shot-aware position embeddings to model inter-shot relationships. CineTrans~\cite{wu2025cinetrans} designs a mask-based control mechanism for shot transitions. 2) \textit{Autoregressive and Interactive Long Video Generation Model}: Baselines include Self Forcing~\cite{huang2025self_forcing}, LongLive~\cite{yang2025longlive}, Rolling Forcing~\cite{liu2025rolling_forcing}, and Infinity-RoPE~\cite{yesiltepe2025infinity}. LongLive~\cite{yang2025longlive} utilizes a KV-recache mechanism to refresh states with new prompts, enabling interactive generation. Infinity-RoPE~\cite{yesiltepe2025infinity} introduces a training-free RoPE Cut for multi-scene transitions in continuous rollouts. Although these autoregressive models can generate videos of several hundred frames, their multi-shot generation capabilities remain limited.

\begin{figure*}[!t]
  \centering
   \includegraphics[width=1.0\linewidth]{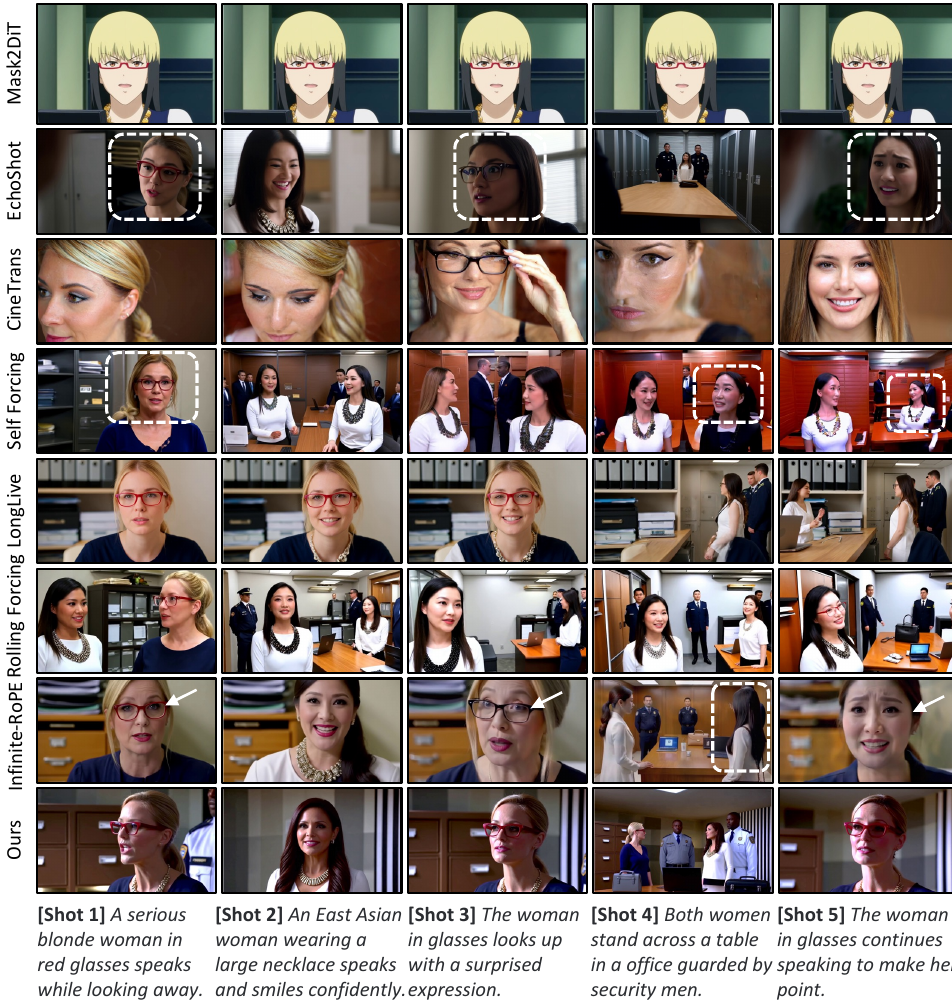}
   \vspace{-15pt}
   \caption{Qualitative Comparison. We present the initial frames of each shot generated by all compared methods. Our approach not only adheres strictly to the prompts and maintains high visual coherence, but also produces natural transitions between shots.} 
   \label{fig:compare}
   \vspace{-6pt}
\end{figure*}

\begin{table*}[!t]
    \centering
    \caption{User Preference Rate. Participants select their preferred video from a randomized set of results generated by all methods. Multiple selections are allowed.}
    \vspace{-7pt}
    \label{table:user_study}
    \renewcommand{\arraystretch}{1.3} 
    % \resizebox{\linewidth}{!}{
        \begin{tabular}{l|cccccccc}
            \toprule
            % \diagbox{左下文字}{右上文字}
            % 这里 "Items" 是纵向列的标题，"Method" 是横向行的标题
            Method & Mask2DiT & EchoShot & CineTrans & Self~Forcing  & LongLive & Rolling~Forcing & Infinity-RoPE& Ours \\
            \midrule
            Visual Consistency & 3.08$\%$ & 12.31$\%$ & 6.21$\%$ & 1.54$\%$ & 12.31$\%$ & 15.38$\%$ & 16.92$\%$ & \textbf{87.69$\%$}\\
            Prompt Following & 0.83$\%$  & 3.08$\%$ & 1.54$\%$ & 10.77$\%$ & 16.15$\%$ & 16.15$\%$ & 14.62$\%$ & \textbf{76.15$\%$}\\
            Visual Quality &  7.69$\%$& 18.46$\%$& 16.92$\%$& 10.77$\%$& 18.46$\%$& 23.08$\%$ & 15.38$\%$ & \textbf{83.08$\%$} \\
            \bottomrule
        \end{tabular}
    % }
% \vspace{-5pt}
\end{table*}

\begin{table*}[!t]
    \centering
    \caption{Ablation Study on the Bidirectional Next-Shot Teacher Model Design.}
    \vspace{-8pt}
    \label{table:ablation_teacher}
    \setlength\tabcolsep{4.5pt} % 适当加大间距
    \renewcommand{\arraystretch}{1.3} 
    % \resizebox{\linewidth}{!}{
        \begin{tabular}{l|cc|ccc|cccc}
            \toprule
            \multirow{2}{*}{Method} & \multicolumn{2}{c|}{Intra-shot Cons.} & \multicolumn{3}{c|}{Inter-shot Cons.} & \multirow{2}{*}{\makecell{Trans. \\ Control $\uparrow$}} & \multirow{2}{*}{\makecell{Text \\ Align. $\uparrow$}} & \multirow{2}{*}{\makecell{Aesthetic \\ Quality $\uparrow$}} & \multirow{2}{*}{\makecell{Dynamic \\ Degrees $\uparrow$}} \\
            \cmidrule(lr){2-3} \cmidrule(lr){4-6}
            & Sub. $\uparrow$ & Bg. $\uparrow$ & Semantic $\uparrow$ & Sub. $\uparrow$ & Bg. $\uparrow$ & & & & \\
            \midrule
            \multicolumn{10}{c}{Context Frames Sampling Strategy} \\
            \midrule
            First Only          & 0.789 & 0.793 & 0.671 & 0.618 & 0.612 & 0.956  & 0.212 & 0.502 & 61.55 \\
            First $\&$ Last     & 0.809 &  \textbf{0.827} & 0.709  & 0.629 & 0.638 & 0.969 & 0.223 & 0.528 &  62.08\\
            Dynamic (Ours)  & \textbf{0.825} & 0.819 &\textbf{0.762} & \textbf{0.654} & \textbf{0.645}& \textbf{0.978} & \textbf{0.234} & \textbf{0.571} & \textbf{63.06} \\
            \midrule
            \multicolumn{10}{c}{Condition Frames Captioning Strategy} \\
            \midrule
            Target Caption      & 0.804 & 0.818 & 0.681 & 0.609 & 0.572 & 0.937 & 0.194 & 0.422 &62.86 \\
            Multi-Captions (Ours)& \textbf{0.825} & \textbf{0.819} &\textbf{0.762} & \textbf{0.654}& \textbf{0.645} & \textbf{0.978} & \textbf{0.234} & \textbf{0.571} & \textbf{63.06} \\
            \midrule
            \multicolumn{10}{c}{Condition Injection Mechanism} \\
            \midrule
            Channel Concat      &  0.814 & 0.802 & 0.743 & 0.628 & 0.608 & 0.912 & 0.223 & 0.509 & 61.43 \\
            Frame Concat (Ours)  & \textbf{0.825} & \textbf{0.819} &\textbf{0.762} & \textbf{0.654} & \textbf{0.645} & \textbf{0.978} & \textbf{0.234} & \textbf{0.571} & \textbf{63.06} \\
            \midrule
            \multicolumn{10}{c}{Training Strategy}\ \\
            \midrule
            Full & 0.816 & 0.810 & 0.749 & 0.631 & 0.624 & 0.969 & 0.227 & 0.546 & 60.85 \\
            Only 3D (Ours)  & \textbf{0.825} & \textbf{0.819} &\textbf{0.762} & \textbf{0.654} & \textbf{0.645} & \textbf{0.978} & \textbf{0.234} & \textbf{0.571} &\textbf{63.06} \\
            \bottomrule
        \end{tabular}
    % }
    \vspace{-2pt}
\end{table*}

\noindent\textbf{Quantitative Results.}
We validate ShotStream using the evaluation set described in Sec.~\ref{sec:experiment_setup}. As shown in Table~\ref{table:quantitative_results}, our model outperforms competing methods across major metrics. It achieves the highest visual consistency while maintaining precise control over shot transitions,  reflected by higher Consistency and Transition Control scores. Additionally, our method demonstrates superior prompt alignment for individual shots and higher overall aesthetic quality. We also evaluate inference efficiency of all methods using a single H200 GPU. Compared to bidirectional models, our approach yields more than 25$\times$ improvement in throughput (FPS). It also enables autoregressive long multi-shot video generation with minimal speed degradation relative to causal long-video models.

\noindent\textbf{Qualitative Results.}
In Fig.~\ref{fig:compare}, we provide a qualitative comparison on a complex, narrative-driven multi-shot prompt to illustrate the superiority of our method. Baseline methods, including Mask2DiT~\cite{qi2025mask2dit}, CineTrans~\cite{wu2025cinetrans}, Self Forcing~\cite{huang2025self_forcing}, and Rolling Forcing~\cite{liu2025rolling_forcing} fail to generate shots that align with their respective prompts. While EchoShot~\cite{wang2025echoshot} and Infinity-RoPE~\cite{yesiltepe2025infinity} successfully adapt to individual shot instructions, they struggle with inter-shot consistency. LongLive~\cite{yang2025longlive} confuses the identities of the two women appearing throughout the sequence. In contrast, our method faithfully adheres to multi-shot prompts while achieving high visual consistency and coherence with smooth transitions.

\subsection{User Study}
Due to the subjective nature of evaluating multi-shot video generation, we conducted a user study to compare different methods and validate the perceptual advantages of our proposed ShotStream. We randomly sampled 24 multi-shot prompts from the evaluation set (detailed in Sec.~\ref{sec:experiment_setup}). During the test, participants are simultaneously presented with eight videos displayed in a randomized order: one generated by our method and seven from the competing baselines. Participants are asked to evaluate the videos from three aspects: Visual Consistency, Prompt Following, and Visual Quality. Multiple selections are allowed for each question. The user study involves 54 participants, and the results in Table~\ref{table:user_study} indicate that our method is consistently preferred by most users.

\subsection{Ablation Studies}
We perform ablation studies to validate the key design choices and training strategies of the bidirectional next-shot teacher and causal student models.

\begin{table*}[!t]
    \centering
    \caption{Ablation Study on the Causal Student Model Design and Training Strategy.}
    \vspace{-6pt}
    \label{table:ablation_student}
    \setlength\tabcolsep{4.5pt}
    \renewcommand{\arraystretch}{1.3} 
    % \resizebox{\linewidth}{!}{
        \begin{tabular}{l|cc|ccc|cccc}
            \toprule
            \multirow{2}{*}{Method} & \multicolumn{2}{c|}{Intra-shot Cons.} & \multicolumn{3}{c|}{Inter-shot Cons.} & \multirow{2}{*}{\makecell{Trans. \\ Control $\uparrow$}} & \multirow{2}{*}{\makecell{Text \\ Align. $\uparrow$}} & \multirow{2}{*}{\makecell{Aesthetic \\ Quality $\uparrow$}} & \multirow{2}{*}{\makecell{Dynamic \\ Degrees $\uparrow$}} \\
            \cmidrule(lr){2-3} \cmidrule(lr){4-6}
            & Sub. $\uparrow$ & Bg. $\uparrow$ & Semantic $\uparrow$ & Sub. $\uparrow$ & Bg. $\uparrow$ & & & & \\
            \midrule
            \multicolumn{10}{c}{Dual-Cache Distinction Strategy} \\
            \midrule
            w/o Indicator     & 0.813 & 0.816 & 0.728 & 0.507 & 0.465 & 0.967 & 0.203 & 0.549 & \textbf{63.09} \\
            Learnable Emb.    & 0.811 & 0.809 & 0.737 & 0.518 & 0.588 & 0.972 & 0.204 & 0.523 & 61.19\\
            RoPE Offset (Ours)& \textbf{0.825}& \textbf{0.819} & \textbf{0.762} & \textbf{0.654}& \textbf{0.645} & \textbf{0.978} & \textbf{0.234} & \textbf{0.571} &63.06  \\
            \midrule
            \multicolumn{10}{c}{Causal Distillation Training Strategy} \\
            \midrule
            Stage 1 Only      &  0.803 & 0.809 & 0.758 & 0.604 & 0.622 & 0.976 & 0.224& 0.568 & 59.66\\
            Stage 2 Only      & 0.819 & 0.814 & 0.704 & 0.583 &  0.547 & 0.975 & 0.218 & 0.467 & 52.86 \\
            Two Stage (Ours)  & \textbf{0.825} & \textbf{0.819} &\textbf{0.762} & \textbf{0.654}& \textbf{0.645} & \textbf{0.978} & \textbf{0.234} & \textbf{0.571} &\textbf{63.06}  \\
            \bottomrule
        \end{tabular}
    % }
    % \vspace{-6pt}
\end{table*}

\begin{figure*}[!t]
  \centering
   \includegraphics[width=1.0\linewidth]{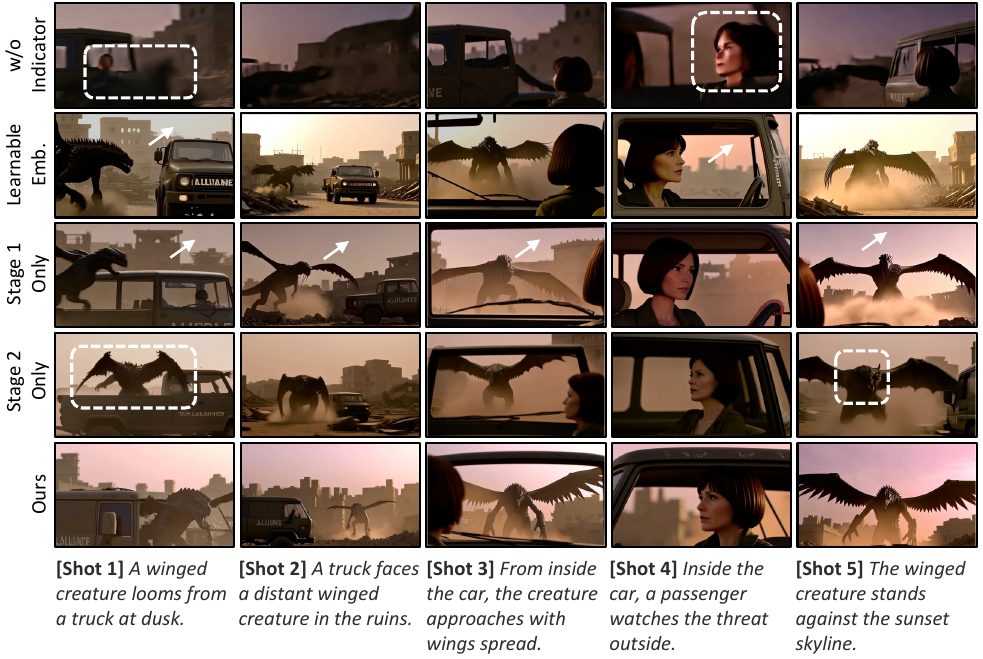}
   \vspace{-12pt}
   \caption{Qualitative Ablation Results for the Causal Student Model. Please refer to \href{https://luo0207.github.io/ShotStream/}{project page} for video comparisons.} 
   \vspace{-8pt}
   \label{fig:ablation}
\end{figure*}

\noindent\textbf{Bidirectional Next-Shot Teacher Model Design.}  As shown in Table~\ref{table:ablation_teacher}, we validate our design choices across four key aspects: 1) \textit{Context Frame Sampling Strategy:} To extract sparse context frames from historical shots, our model adopts a dynamic sampling strategy based on the number of historical shots to fulfill the conditional context budget. This aims to maximize the retention of historical information within a constrained budget. This dynamic approach outperforms naive baselines, such as sampling only the first frame or the first and last frames of each historical shot. 2) \textit{Condition Frames Captioning Strategy}: To validate the necessity of injecting specific prompts for historical shots, we compare our approach against the widely used baseline~\cite{ju2025fulldit,luo2025camclonemaster,ye2025unic} of using the same caption for both condition and target frames. Results indicate that injecting the corresponding prompts for condition frames is necessary and benefits in-context learning by effectively binding historical visual content with its respective text. 3) \textit{Condition Injection Mechanism}: We evaluate our approach of injecting condition frames via temporal token concatenation against the commonly used channel concatenation method~\cite{shin2025motionstream}. This proves superior by allowing the 3D self-attention layers to directly model the relationships between condition and target tokens. 4) \textit{Training Strategy}: We demonstrate that fine-tuning only the 3D self-attention layers outperforms full-parameter fine-tuning. 

\noindent\textbf{Causal Student Model Design.} Table~\ref{table:ablation_student} presents ablations on the model design and distillation strategy of our causal model. 1) \textit{Dual-Cache Distinction Strategy:} To justify the need to separate global from local caches, we compare our proposed RoPE offset (Row 3) against a baseline with no distinction (Row 1) and a variant using a learnable embedding applied to the target video's first chunk (Row 2). The results demonstrate that explicit distinction is essential (Row 1 vs. Row 3), and our training-free RoPE offset outperforms the learnable embedding approach (Row 2 vs. Row 3). 2) \textit{Causal Distillation Training:} We evaluate our two-stage distillation strategy against single-stage baselines (Rows 4 and 5). Both stages prove indispensable: stage 1 establishes foundational next-shot generation capabilities, while stage 2 faithfully simulates inference to bridge the train-test gap. Furthermore, the qualitative results in Fig.~\ref{fig:ablation} reinforce the necessity of both the RoPE offset and the two-stage distillation. Notably, the inter-shot self-forcing distillation significantly improves long-term consistency in visual style and color across the video (Stage 1 Only vs. Ours).

\section{Conclusion}
In this paper, we introduce ShotStream, a novel causal multi-shot video generation architecture that enables interactive long narrative storytelling while achieving 16 FPS on a single GPU. Our core contributions include reformulating the next-shot generation task for streaming, training a bidirectional next-shot teacher model, and distilling it into a causal architecture via a proposed two-stage distillation strategy. Additionally, we introduce a novel dual-cache memory mechanism to ensure visual consistency. Compared to existing bidirectional multi-shot models, ShotStream significantly reduces generation latency and supports streaming prompt inputs at runtime. This empowers users to interactively guide the narrative, adapting upcoming shots based on previously generated content. Furthermore, ShotStream advances the capabilities of autoregressive long video generation models by extending their ability to generate cohesive multi-shot sequences, paving the way for real-time, interactive, long-form storytelling.

\noindent \textbf{Limitations and Future Work.}
While ShotStream is effective for autoregressive multi-shot video generation, we identify two primary limitations. First, we observe visual artifacts and inconsistencies when scenes and text prompts are highly complex. This primarily stems from the limited capacity of our backbone; because our current models are relatively small, we expect that scaling up the base model will improve performance and stability in challenging scenarios. Second, although our method is efficient, it still has room for acceleration to provide better interactive experiences. Techniques such as sparse attention~\cite{jia2025moga,zhuang2025flashvsr} and attention sink~\cite{yang2025longlive,shin2025motionstream} could be integrated into our model to achieve faster generation. We leave these extensions to future research.
{
    \small
    \bibliographystyle{ieeenat_fullname}
    \bibliography{main}
}

\clearpage
\setcounter{page}{1}
\maketitlesupplementary

In this supplementary file, we provide the following materials:
\begin{itemize}
    \item Text-to-Video base model.
    \item Distribution Matching Distillation objective function.
    \item Training details.
\end{itemize}

\section{Text-to-Video Base Model}
\label{sec:supp_t2v}

Our proposed model, ShotStream, builds upon the transformer-based latent diffusion architecture, Wan2.1-T2V-1.3B~\cite{wan2025wan}. The architecture integrates a 3D Variational Auto-Encoder (VAE)~\cite{kingma2013vae} for latent feature mapping alongside a sequence of transformer blocks responsible for modeling temporal dynamics. Each basic transformer block is composed of $3$D spatial-temporal attention, cross-attention, and feed-forward network (FFN). The input text is encoded by the umT$5$ encoder $\varepsilon_{\textrm{umT5}}$~\cite{chung2023umt5} and injected into the architecture through cross-attention layers. The model utilizes the Rectified Flow~\cite{liu2022flowstraightfastlearning} framework to train the diffusion transformer, such that we can generate a data sample $\boldsymbol{x}$ from a starting Gaussian sample $\boldsymbol{z}\in \mathcal{N}(\boldsymbol{0},\boldsymbol{I})$. Specifically, for a data point $\boldsymbol{x}$, its noised version $\boldsymbol{x}_t$ at timestep $t$ is constructed as $\boldsymbol{x}_t = (1 - t)\boldsymbol{x} + t\boldsymbol{z}.$ The training objective is a simple MSE loss:
\begin{equation}
\label{eq:denoising_loss}
\mathcal{L}_{RF}(\theta)= \mathbb{E}_{t,\boldsymbol{x},\boldsymbol{z}}\left \| \boldsymbol{v}_{\theta}(\boldsymbol{x}_t, t, \boldsymbol{c}_{\textrm{text}}) - (\boldsymbol{z}-\boldsymbol{x}) \right \|_2^2,
\end{equation}
where the velocity $\boldsymbol{v}_\theta$ is parameterized by a network $\theta$.

\section{Distribution Matching Distillation Objective}
\label{sec:supp_dmd}
As discussed in the main text, DMD minimizes the reverse KL divergence between the smoothed data distribution and the student generator's output distribution. The gradient for this optimization is approximated by the difference between two score functions: 

\begin{align}
\label{eq:dmd}
\nabla_{\phi} \mathcal{L}_{\text{DMD}} & \triangleq \mathbb{E}_{t} \Big( \nabla_{\phi} \text{KL} \big( \boldsymbol{p}_{\text{gen}, t} \| \boldsymbol{p}_{\text{data}, t} \big) \Big) \notag \\
& \approx - \mathbb{E}_{t} \Bigg( \int \bigg( \boldsymbol{s}_{\text{data}} \Big( \Psi \big( G_{\phi}(\epsilon), t \big), t \Big) \notag \\ 
& \qquad - s_{\text{gen},\xi} \Big( \Psi \big( G_{\phi}(\epsilon), t \big), t \Big) \bigg) \frac{d G_{\phi}(\epsilon)}{d \phi} \, d\epsilon \Bigg),
\end{align}

where $\Psi$ represents the forward diffusion process, $\epsilon$ is random Gaussian noise, $G_{\phi}$ is the student generator parameterized by $\phi$, and $s_{\text{data}}$ and $s_{\text{gen},\xi}$ denote the score functions trained on the data and the student generator's output distributions, respectively, using a denoising loss (Eq.~\ref{eq:denoising_loss}).

\section{Training Details}
\label{sec:supp_training_details}

\noindent\textbf{Dataset and Preprocessing.} We utilize an internally curated dataset of 320K multi-shot videos, with each video comprising 2 to 5 shots and totaling up to 250 frames. Each sample is annotated with hierarchical prompts: a global caption describing the narrative arc, characters, and visual style, and shot-level captions detailing specific actions and content within each segment.

\noindent\textbf{Bidirectional Teacher Training.} We optimize only the 3D self-attention layers within the DiT blocks. Training is conducted for 10,000 steps using the Adam optimizer with a learning rate of $1e-5$ and a batch size of 64.

\noindent\textbf{Causal Adaptation Initialization.} Following the CausVid protocol~\cite{yin2024causalvid}, we initialize the student model with teacher weights and adapt it to a causal attention architecture. We sample 5K ODE solution pairs from the teacher and train all model parameters for 2,000 steps with a learning rate of $1e-6$ and a batch size of 64. This alignment on ODE trajectories bridges the gap between the bidirectional teacher and causal student, stabilizing subsequent distillation.

\noindent\textbf{Causal Distillation.} Distillation proceeds in two stages:
\begin{itemize}
    \item Stage 1: Intra-shot Self-forcing. This stage employs ground-truth historical shots from our dataset. The model is trained to predict the immediate next shot conditioned on perfect history. This stage converges quickly around 500 steps with a batch size of 32. Following DMD2 \cite{yin2024dmd2}, we set the learning rates to $2e-6$ for the generator and $4e-7$ for the critic, with a 1:5 update ratio.
    \item Stage 2: Inter-shot Self-forcing. Using a 5-shot subset of our dataset, the model is trained on its own multi-shot rollouts. For each iteration, the model generates a sequence of 5-second shots; when a shot boundary is reached, the global context cache is updated with the generated content while the local cache is reset. We apply LoRA tuning for 1,000 steps using the same learning rates as Stage 1.
\end{itemize}
All experiments are conducted on a cluster of 32 NVIDIA H800 GPUs.

\end{document}